\documentclass[10pt,twocolumn,letterpaper]{article}
\usepackage[pagenumbers]{cvpr}
\usepackage{amsmath,amsfonts,bm}

\def\eqref#1{equation~\ref{#1}}

\def\1{\bm{1}}

\DeclareMathAlphabet{\mathsfit}{\encodingdefault}{\sfdefault}{m}{sl}
\SetMathAlphabet{\mathsfit}{bold}{\encodingdefault}{\sfdefault}{bx}{n}

\usepackage{marvosym}

\makeatletter
\patchcmd{\maketitle}
  {\renewcommand\thefootnote{\@fnsymbol\c@footnote}}
  {\renewcommand\thefootnote{\Letter}}
  {}
  {\PackageError{dataloop}{Could not set the corresponding-author marker}{}}
\patchcmd{\@maketitle}
  {\vspace*{24pt}}
  {\vspace*{12pt}}
  {}
  {\PackageError{dataloop}{Could not adjust the title-to-author spacing}{}}
\makeatother

\usepackage{multirow}
\usepackage{array}
\usepackage{tabularx}
\usepackage[normalem]{ulem}
\usepackage{ragged2e}
\usepackage{needspace}
\newcolumntype{Y}{>{\RaggedRight\arraybackslash}X}

\usepackage{tcolorbox}
\tcbuselibrary{skins}
\newcounter{algo}
\newtcolorbox{algobox}[1]{%
  enhanced, sharp corners=downhill,
  colback=white, colframe=black!70, boxrule=0.5pt,
  left=8pt, right=8pt, top=6pt, bottom=6pt,
  attach boxed title to top left={xshift=8pt, yshift=-8pt},
  boxed title style={colback=black!8, colframe=black!70, boxrule=0.5pt,
    sharp corners, size=small, top=1pt, bottom=1pt, left=4pt, right=4pt},
  coltitle=black, fonttitle=\normalsize,
  title={#1}}

\newcommand{\basemodel}{Qwen3.5-0.8B}
\newcommand{\teachermodel}{Qwen3.8-27B}

\newcommand{\bestAvg}{63.04\%}
\newcommand{\bestRound}{R33}
\newcommand{\gainAvg}{5.32\,pp}          %

\definecolor{cvprblue}{rgb}{0.21,0.49,0.74}
\usepackage[pagebackref,breaklinks,colorlinks,allcolors=cvprblue]{hyperref}
\def\confName{CVPR}
\def\confYear{2027}

\title{MM-DataLoop: Closing the Data-Side Loop for Recursive Self-Improvement in MLLM Post-Training}

\author{%
Likang Peng \quad Chen Ju\thanks{Corresponding Author.} \quad Xu Chen \quad Xiangyang Wu\\
Jinsong Lan \quad Haihong Tang \quad Ying Chen\\
Alibaba Group\\
{\small\strut}%
}

\hypersetup{%
  pdftitle={MM-DataLoop: Closing the Data-Side Loop for Recursive Self-Improvement in MLLM Post-Training},
  pdfauthor={Likang Peng, Chen Ju, Xu Chen, Xiangyang Wu, Jinsong Lan, Haihong Tang, Ying Chen}%
}

\begin{document}
\maketitle
\begin{abstract}
Recursive self-improvement (RSI) offers a path toward improving models through repeated evaluation and revision. We study its data side in multimodal language model (MLLM) post-training, where heterogeneous operations and cross-round evidence must be coordinated. MM-DataLoop organizes data interventions into five operation families and connects them through an Observe--Diagnose--Plan--Execute--Verify--Reflect cycle with shared rule memory. Across 34 candidate rounds on Qwen3.5-0.8B and five benchmarks, we hold the base initialization, shared training schedule and evaluation protocol fixed while varying data recipes. The five-benchmark mean reaches 63.04 at R33, up 5.32 points from the zero-shot baseline; R29, R32, and R33 form a close top cluster within 0.09 points. In eight-candidate comparisons, guided search reaches the preselected target in six candidates and identifies the top observed candidate, while random arms show higher means and lower dispersion. The selected R29 recipe also scores above two full-pool baselines under the same epoch schedule. Together, these findings establish MM-DataLoop as a traceable workflow and provide an empirical basis for studying how data-side decisions shape outcomes across MLLM post-training recipes.
\end{abstract}

\section{Introduction}
\label{sec:intro}

Recursive self-improvement (RSI) relies on repeated evaluation and revision. In MLLM post-training, this process can also operate on the training data: benchmark outcomes inform subsequent acquisition, synthesis, denoising, and recipe composition. While much work focuses on model updates or autonomous experiment design, coordinating heterogeneous data operations and carrying evidence from one recipe to the next remains central to the data side of the loop. We study MM-DataLoop as a workflow for doing so with benchmark feedback and recorded experience.

Recent curation work studies selection and data quality~\citep{gadre2023datacomp,penedo2024fineweb,xie2023dsir}, mixture optimization~\citep{xie2023doremi}, and fixed-protocol data-selection comparisons~\citep{shin2026dcvlr}. CurateEvo evolves curation code from failed agent trajectories for agentic post-training~\citep{wang2026curateevo}; MMPostTrainBench evaluates autonomous multimodal post-training research with development feedback and independent evaluation~\citep{liu2026mmposttrainbench}. MM-DataLoop addresses the data-preparation workflow itself: it links acquisition, synthesis (including CoT production), denoising, and mixture composition to model feedback and persistent operational rules. Its 34-round record makes these transformations and their benchmark profiles inspectable across candidate recipes.

\begin{figure*}[t]
\centering
\includegraphics[width=0.83\linewidth]{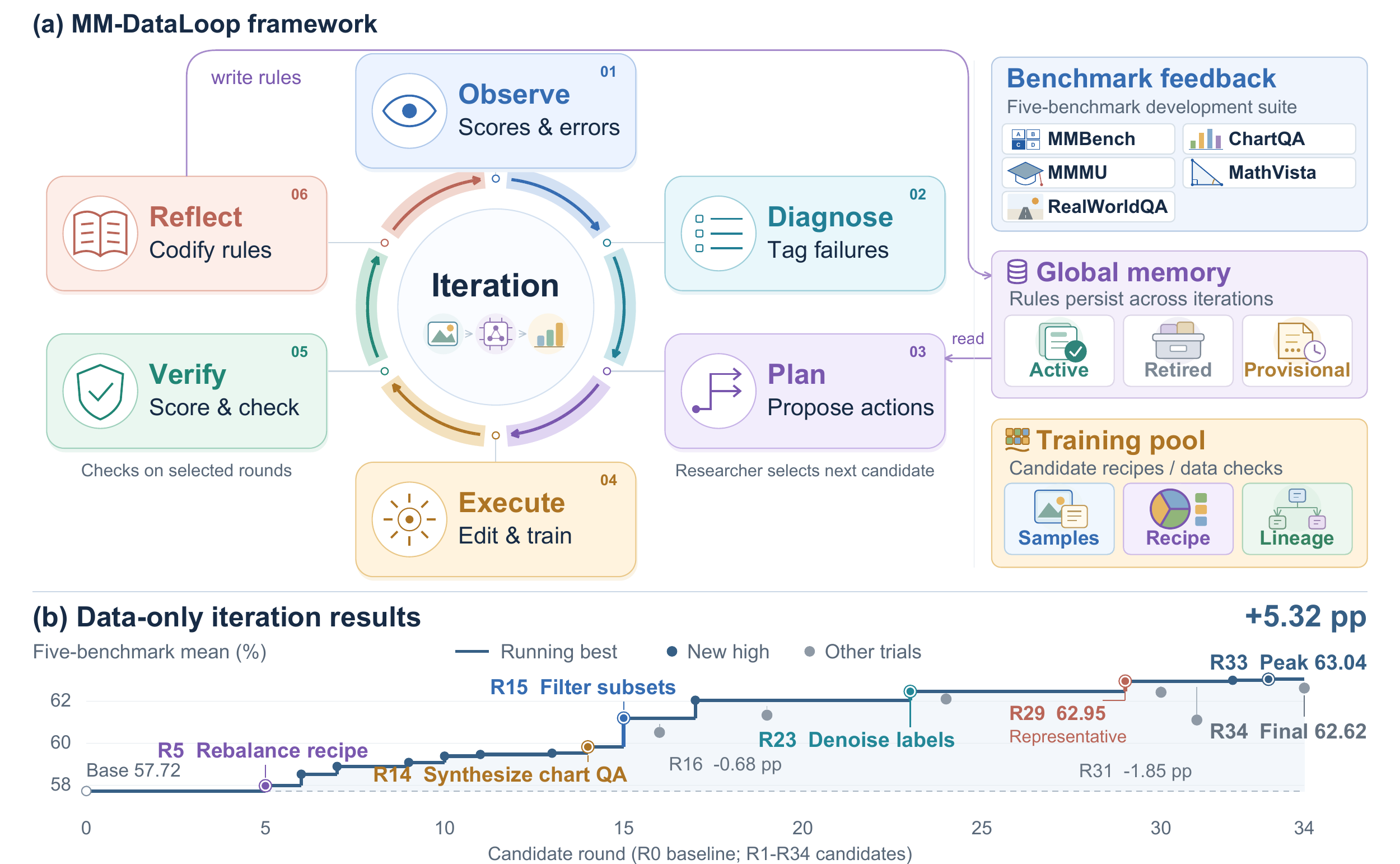}
\caption{MM-DataLoop overview. (a) The six-stage data-side loop connects frozen benchmark feedback, scoped rule memory, and an evolving training pool. (b) The five-benchmark trajectory summarizes candidate outcomes and key data interventions.}
\label{fig:mm-dataloop}
\end{figure*}

We present MM-DataLoop, a data-side iteration workflow for MLLM post-training, organized around five operation families---Acquire, Synthesize, Denoise, Compose, and Codify. Its six stages---Observe, Diagnose, Plan, Execute, Verify, and Reflect---carry benchmark feedback and prior outcomes through a shared rule memory. In our empirical study, candidates share a base initialization and training schedule while data recipes change, creating a consistent protocol for comparing outcomes. As recipe size changes, the fixed-epoch schedule also changes training exposure; we report pool size and optimizer steps alongside scores. A five-tag diagnostic vocabulary structures feedback to planning, while the experiment log also retains unclassified anomalies for subsequent review.

Across 34 candidate rounds on \basemodel{}, the highest reported equal-weight mean on MMBench~\citep{liu2024mmbench}, ChartQA~\citep{masry2022chartqa}, MMMU~\citep{yue2024mmmu}, MathVista~\citep{lu2024mathvista}, and RealWorldQA~\citep{realworldqa2024} is \bestAvg{} at \bestRound{}, a \gainAvg{} increase over the zero-shot baseline. R29, R32, and R33 form a closely spaced top cluster within 0.09 points. In an eight-candidate comparison, guided search reaches the preselected target in six candidates, while random-search arms yield higher means and lower dispersion. The results reveal how targeted proposals and broad sampling expose different useful regions of the data-operation space. Our contributions are:

\noindent $\bullet$ \textbf{An end-to-end data-operation framework.} Five operation families and a six-stage cycle connect data acquisition, example production and transformation, filtering, recipe composition, and cross-round rule memory.

\noindent $\bullet$ \textbf{Empirical evidence across data interventions.} The 34-round record links synthesis, CoT supervision, source and subset filtering, answer-format choices, and image-resolution changes to benchmark-level outcomes under a consistent training protocol.

\noindent $\bullet$ \textbf{Comparative reference points for curation.} Curated-versus-full-pool, selective-versus-random deletion, and guided-versus-random search comparisons provide complementary evidence on recipe quality, target attainment, and score consistency.

\section{Related Work}
\label{sec:related}

\paragraph{Model-Side Recursive Self-Improvement.}
Recent RSI research includes methods that iterate the model or policy using its own outputs. Examples include self-rewarding training~\citep{yuan2024selfrewarding}, self-play~\citep{chen2024spin}, and reinforced self-training~\citep{gulcehre2023reinforced,singh2023beyond}. Rationale distillation~\citep{zelikman2022star,yuan2023scaling} and constitutional self-critique~\citep{bai2022constitutional} also use generated feedback or supervision. Reflexion uses linguistic feedback and episodic memory to improve agent performance across trials~\citep{shinn2023reflexion}, while the AI Scientist organizes research into an autonomous experiment loop~\citep{lu2024aiscientist}. MM-DataLoop focuses this iteration on data, holding the base-model initialization, training hyperparameters, and evaluation protocol fixed.

\paragraph{MLLM Post-Training \& Data Recipes.}
Multimodal applications include retrieval~\citep{chen2026pailitao,cheng2023category,cheng2023mixer,wu2026triad}, video understanding~\citep{ju2022prompting}, virtual try-on~\citep{chen2024wear}, human animation and background generation~\citep{liu2026improving,yao2025beyond}, and audio-visual localization and segmentation~\citep{liu2022exploiting,liu2023audio,liu2024audio,liu2024annotation}. Inference acceleration~\citep{ju2024turbo,wang2025folder} complements the data and training stack. MLLM post-training uses instruction corpora alongside efficient adaptation methods such as LoRA~\citep{hu2022lora}. Data collections range from general instruction corpora~\citep{liu2023llava,liu2024llava15,chen2024sharegpt4v} to domain-specific pools~\citep{shi2024mathllava,masry2022chartqa}. More recent systems combine broader collections with stronger backbones~\citep{chen2024internvl,wang2024qwen2vl,bai2025qwen25vl,tong2024cambrian}. Recipe studies compare data mixtures and their effects on model performance~\citep{mckinzie2024mm1,laurencon2024matters,deitke2024molmo,li2024llavaonevision,shi2024eagle}. Related work studies contrastive pre-training and embeddings~\citep{wang2025advancing,wang2025explore}, latent reasoning supervision~\citep{li2026structural}, temporal localization and grounding under varied supervision~\citep{zhao2020bottom,ju2021divide,ju2022adaptive,ju2023distilling,ju2023multi,ju2023constraint,wang2025contrast}, and open-vocabulary action recognition~\citep{ju2026r}. MM-DataLoop records how each result informs later data revisions, supporting continued search.

\paragraph{Data Curation, Agent Memory, \& Evaluation.}
Data curation studies selection~\citep{gadre2023datacomp,penedo2024fineweb,xie2023dsir}, mixture optimization~\citep{xie2023doremi,ye2024mixinglaws}, and processing pipelines~\citep{chen2024datajuicer}. Token-level governance refines individual samples~\citep{lin2026squeeze}; complementary visual learning methods study unified tokenization~\citep{chen2025wave} and open-vocabulary or annotation-light segmentation~\citep{ma2023attrseg,ma2023diffusionseg,ma2025freesegdiff,yang2024multi}. DataEnvGym~\citep{khan2025dataenvgym} formalizes modular teacher-agent environments that map student errors or weak skills to targeted generation plans and evaluate the resulting examples through updated-student feedback. Its focus is teacher-agent policies. MM-DataLoop instead connects acquisition, synthesis and rewriting, denoising, mixture composition, and rule codification through persistent outcome records. It evaluates complete training-pool recipes from a shared base checkpoint with fresh adapters over a 34-round MLLM trajectory, using full-pool, random-deletion, and random-search reference arms to inspect interactions between operations.

Table~\ref{tab:dataenvgym-scope} makes the comparison concrete at the level of experimental object and evidence.

\begin{table*}[t]
\centering
\caption{Complementary scopes of DataEnvGym and MM-DataLoop. The comparison focuses on the unit each study organizes and evaluates.}
\label{tab:dataenvgym-scope}
\fontsize{8pt}{9.6pt}\selectfont
\setlength{\tabcolsep}{6pt}
\begin{tabularx}{\textwidth}{@{}l Y Y@{}}
\toprule
Study dimension & DataEnvGym~\citep{khan2025dataenvgym} & MM-DataLoop \\
\midrule
Iteration unit & Teacher-agent policy and its data-generation actions & Training-pool recipe across candidate rounds \\
Operation scope & Targeted example generation in modular teacher environments & Acquisition, synthesis, filtering/denoising, and mixture composition \\
Training contract & Student feedback within the teacher-agent environment & Fresh LoRA adapter from a shared base for every recipe \\
Empirical record & Modular policy evaluation across environment configurations & 34-round MLLM recipe trajectory with random-search and random-deletion arms \\
\bottomrule
\end{tabularx}
\end{table*}

Complementary systems study fixed-protocol selection of multimodal reasoning data~\citep{shin2026dcvlr}, evolve executable curation strategies from failed agent trajectories~\citep{wang2026curateevo}, and evaluate autonomous multimodal post-training with development feedback and independent evaluation~\citep{liu2026mmposttrainbench}. MM-DataLoop analyzes interacting curation operations over an evolving multimodal training pool, preserving benchmark profiles for cross-round recipe comparisons.

\section{MM-DataLoop}
\label{sec:system}

\subsection{Overview}
\label{sec:overview}

MM-DataLoop couples benchmark-driven diagnosis, a five-family data-operation space, and scoped rule memory in a recurring optimization of the training-pool recipe. The workflow links candidate outcomes to later interventions, while the fresh-base training contract keeps the model initialization fixed across rounds. Figure~\ref{fig:mm-dataloop} summarizes these interacting components.

\subsection{Data Operation Families}
\label{sec:families}

Data operations fall into five families according to what they change. The classification applies to the operation rather than the tool: generating samples is Synthesize, while judging or removing existing samples is Denoise. Each supported operation has one family assignment.

\textbf{\emph{Acquire}} brings external assets into the pipeline through ingestion, extraction, or OCR, with source and preprocessing information. \textbf{\emph{Synthesize}} creates new examples from existing assets, including rewritten QA, CoT demonstrations, and MCQs; provenance fields cover the source model, prompt, and verification where recorded. \textbf{\emph{Denoise}} repairs, normalizes, or removes existing samples through rules or model judgments.

\textbf{\emph{Compose}} changes collection-level properties such as subset ratios, length distribution, difficulty, and curriculum to form candidate recipes. \textbf{\emph{Codify}} turns experience into structured rules for global memory. Table~\ref{tab:ops} lists the operation types represented in the framework; each candidate recipe specifies the operations used in that round.
\label{sec:ops}

\begin{table*}[t]
\centering
\caption{Overview of Data Operations. In I/O, $X$ = sample, $Q$/$A$ = question/answer, $T$ = teacher.}
\label{tab:ops}
\fontsize{8pt}{9.6pt}\selectfont
\setlength{\tabcolsep}{10pt}
\begin{tabular}{@{}l c l l@{}}
\toprule
Operation & Family & Input $\rightarrow$ Output & Main Use \\
\midrule
Dataset Ingestion     & Acquire    & URL $\rightarrow$ Asset                 & Source Selection, License, Versioning \\
Source Segmentation   & Acquire    & Doc $\rightarrow$ Units                 & Segmentation, OCR, Frame Extraction \\
Teacher QA Rewriting  & Synthesize & $(Q,A) \rightarrow (Q',A')$             & Form / Style / Length Alignment \\
CoT Distillation      & Synthesize & $(Q,T) \rightarrow (Q,\text{CoT}+A)$    & Rejection-Sampled CoT Distillation \\
MCQ Conversion        & Synthesize & $(Q,A) \rightarrow (Q,\text{opts})$     & Free-Form $\rightarrow$ MCQ \\
Instruction Evolution & Synthesize & $Q \rightarrow Q'$                      & Difficulty / Diversity Expansion \\
Rule-based Filtering  & Denoise    & $X \rightarrow \text{Keep/Drop}$        & Length / Language / Corrupt-Image \\
Structural Repair     & Denoise    & $X \rightarrow X'$             & \texttt{<image>} Repair, Format Align \\
Verifier Screening    & Denoise    & $(X,T) \rightarrow \text{Keep/Drop}$    & Teacher-as-Verifier \\
Mixture Weighting     & Compose    & Subsets $\rightarrow$ Recipe            & Static Subset Weighting \\
Curriculum Design     & Compose    & Pool $\rightarrow$ Ordered Recipe       & Staged / Leveled Schedule \\
Rule Codification     & Codify     & Obs.\ $\rightarrow$ Rule Entry          & Experience $\rightarrow$ Memory \\
\bottomrule
\end{tabular}
\end{table*}

\subsection{Data Iteration Cycle}
\label{sec:cycle}

The workflow is expressed as Observe~$\to$~Diagnose~$\to$~Plan~$\to$~Execute~$\to$~Verify~$\to$~Reflect. A candidate round produces model $\hat F_t$ and pool $\hat P_t$; its record includes the recipe and benchmark outcome. R0 denotes the original base checkpoint $F_0$, and $B$ the five-benchmark development suite. The next recipe uses available results and project state, connecting candidate outcomes to subsequent data decisions in the next round.

\textbf{\emph{Observe}} evaluates a model and records scores and errors. \textbf{\emph{Diagnose}} organizes candidate failure modes (\S\ref{sec:diagnose}). \textbf{\emph{Plan}} uses available tags, prior rules, scores, and experiment history to propose hypotheses $H_t$ and operations $\Pi_t$. \textbf{\emph{Execute}} builds a candidate pool and trains a fresh LoRA adapter from $F_0$. \textbf{\emph{Verify}} evaluates candidate outcomes and applies the structural checks available for the candidate data. \textbf{\emph{Reflect}} records the outcome and any resulting rule for later consideration. Together, these stages organize the documented cycle from benchmark feedback to recipe revision.

The workflow combines scripts for evaluation and data operations with LLM-assisted diagnosis and planning. Available client code identifies Qwen3.8-27B as the configured default synthesis model. The analysis uses a five-tag failure vocabulary and a bounded operation menu; researcher decisions, benchmark feedback, and LLM-assisted proposals jointly shape candidate recipes across the recorded search.

Algorithm~\ref{alg:round} gives the round update.

\begin{figure*}[t]
\refstepcounter{algo}\label{alg:round}
\begin{algobox}{Algorithm~\thealgo: One Round of MM-DataLoop}
\small
\noindent\textbf{Input:} original base $F_0$; retained state $(F_{t-1},P_{t-1},M_{t-1})$; benchmarks $B$; failure vocabulary $V$; experiment history $\mathcal{H}_{t-1}$.\\
\textbf{Output:} retained state $(F_t,P_t,M_t)$ and history $\mathcal{H}_t$.\\
\textbf{Notation:} $S_t^{\mathrm{obs}},E_t$ = observed scores and errors; $S_t$ = candidate scores; $D_t$ = tags; $R_t$ = retrieved rules; $H_t,\Pi_t$ = hypotheses and operations; $J_t$ = candidate record.
\vspace{2pt}
\hrule
\vspace{4pt}
\renewcommand{\arraystretch}{1.15}
\begin{tabular}{@{}r@{\ \ }l@{\quad}l@{}}
1. & $(S_t^{\mathrm{obs}},E_t)\gets\textsc{Evaluate}(F_{t-1},B)$ & $\triangleright$ Observe \\
2. & $D_t\gets\textsc{Diagnose}(E_t;V)$ & $\triangleright$ Diagnose \\
3. & $R_t\gets\textsc{Retrieve}(M_{t-1};D_t,B)$ & $\triangleright$ Active rules in scope \\
4. & $(H_t,\Pi_t)\gets\textsc{Plan}(D_t,R_t,S_t^{\mathrm{obs}},\mathcal{H}_{t-1})$ & $\triangleright$ Plan \\
5. & $\hat P_t\gets\textsc{Execute}(P_{t-1},\Pi_t);\quad\hat F_t\gets\textsc{Train}(F_0,\hat P_t)$ & $\triangleright$ Fresh adapter \\
6. & $J_t\gets\textsc{Verify}(\hat F_t,\hat P_t,B);\quad S_t\gets J_t.\mathrm{scores}$ & $\triangleright$ Candidate results \\
7. & $(F_t,P_t)\gets\textsc{SelectNext}(F_{t-1},P_{t-1},\hat F_t,\hat P_t,J_t)$ & $\triangleright$ next state, informed by recorded results \\
8. & $\mathcal{H}_t\gets\mathcal{H}_{t-1}\cup\{(t,H_t,\Pi_t,\hat P_t,J_t)\}$ & $\triangleright$ retain candidate record \\
9. & $M_t\gets\textsc{Codify}(M_{t-1},H_t,J_t)$ & $\triangleright$ Reflect for later planning \\
\end{tabular}
\end{algobox}
\end{figure*}

\subsection{Failure Diagnostic Vocabulary}
\label{sec:diagnose}

Diagnose uses three sample-level tags and two aggregate tags. \emph{\uline{Format-violation}} marks outputs that violate the required answer format. \emph{\uline{Hallucination}} marks content unsupported by visual or factual evidence. \emph{\uline{Reasoning-skip}} marks a missing intermediate step when an explicit reasoning trace is required. At the aggregate level, \emph{\uline{length-drift}} marks a shift in answer-length distribution, and \emph{\uline{task-mix-mismatch}} marks task-level changes associated with the data mixture. Tags may co-occur. Errors not covered by the vocabulary remain available as additional anomalies for later review.

\subsection{Global Memory Schema}
\label{sec:memory}

Global memory stores rules and retains failure records across rounds. Each rule carries the fields in Table~\ref{tab:memory-fields}. Reflect writes rules through Codify; Plan retrieves entries whose scope matches the current task and whose status is \emph{active}. Codify can create, update, or retire a rule. Retired entries remain available for audit but are excluded from default planning reads under the active-rule policy.

For example, a rewriting candidate with a lower benchmark score can motivate a rule favoring verification-based filtering. The entry records its supporting rounds and applicable data settings. A later plan can retrieve that rule when considering label-denoise. New evidence can narrow a rule's scope or retire it; its supporting record remains available to inform subsequent planning and audits.

\Needspace{12\baselineskip}
\subsection{Candidate Evaluation and Data Checks}
\label{sec:verify}

Candidate outcomes are recorded across five benchmarks, and recipe comparisons inform subsequent search. Structural preflight validation is available in the data pipeline and was used for selected candidate datasets, including the later random-search arms. Candidate selection integrates LLM-assisted proposals, scripted data operations, benchmark results, and researcher decisions. Each iteration therefore links its recipe to the corresponding measured benchmark profile for direct comparison.

\begin{table}[t]
\centering
\caption{Global Memory Fields. Entries link operational recommendations to their evidence and scope.}
\label{tab:memory-fields}
\fontsize{8pt}{9.6pt}\selectfont
\setlength{\tabcolsep}{5pt}
\begin{tabularx}{0.88\linewidth}{@{}l Y@{}}
\toprule
Field & Semantics \\
\midrule
\emph{text}     & natural-language rule with an explicit applicability condition \\
\emph{evidence} & rounds supporting or refuting the rule; positive and negative evidence share this set \\
\emph{scope}    & applicability tags: benchmark, task form, data-operation family, or model configuration \\
\emph{status}   & \emph{active} / \emph{retired} / \emph{provisional}; only \emph{active} rules participate in default planning reads \\
\emph{hint}     & operational hint to next-round planning, mappable to operations or families of Table~\ref{tab:ops} \\
\emph{lineage}  & creation/update context: round index, triggering hypothesis, verification-snapshot pointer \\
\bottomrule
\end{tabularx}
\end{table}

\begingroup
\let\makeLineNumberOdd\makeLineNumberRight
\let\makeLineNumberEven\makeLineNumberRight
\section{Iteration Study Setup}
\endgroup
\label{sec:setup}

We report 34 candidate rounds on five multimodal benchmarks. The base-model initialization, optimizer, batch, epoch, and evaluation settings are shared across candidates, while data recipes vary. The same five-benchmark suite provides a consistent development/evaluation signal throughout diagnosis, selection, and planning. Accordingly, scores characterize outcomes within this iterative study.

\paragraph{Base Model \& Training Protocol.}
\label{sec:basemodel}
The base model is \basemodel{} (instruct). Each round starts from the same base checkpoint with a newly initialized LoRA adapter, keeping learned weights independent across recipes. We use LoRA ($r{=}32$, $\alpha{=}64$), AdamW, a cosine schedule, effective batch size 128, three epochs, and cutoff length 2{,}048. The configured synthesis teacher is \teachermodel{}; available client code records this default model identifier. The starting pool draws from LLaVA~\citep{liu2023llava,liu2024llava15}, ShareGPT4V~\citep{chen2024sharegpt4v}, MathV360K~\citep{shi2024mathllava}, and ChartQA~\citep{masry2022chartqa}. Because pool size changes across rounds, the fixed three-epoch schedule yields recipe-dependent numbers of examples, tokens, and optimizer steps; we report available step counts with the results.

\paragraph{Evaluation Protocol and Use.}
\label{sec:eval}
Evaluation covers the five benchmarks in Table~\ref{tab:decomp}, including MCQ and free-form tasks. The four-benchmark evaluation package uses temperature 0.7, top-$p$ 0.8, top-$k$ 20, and presence penalty 1.5; MMMU uses its separate kit at temperature 0, top-$p$ 1, and top-$k$ $-1$. We use the available fixed prompts, post-processing, and rule-based answer matching, yielding a consistent protocol for within-study comparisons. Absolute scores may differ from official model reports; throughout the search, the benchmark suite supplies development feedback for comparing candidate recipes.

\paragraph{Scoring \& Seed Protocol.}
\label{sec:scores}
$S_{t,b}$ is candidate round $t$'s accuracy on benchmark $b$, in percent; $\bar S_t$ is the equal-weight five-benchmark mean. $\Delta$ R0 compares with the zero-shot baseline, while $\Delta$ Prev compares with the preceding numbered round. The four-benchmark suite averages seeds 42 and 43; MMMU uses two evaluation runs. Candidate means use the available complete benchmark scores. R2, R3, and R23 also have additional same-checkpoint evaluation records for repeatability checks.

\paragraph{Compute Budget.}
\label{sec:compute}
All candidates use the shared three-epoch training schedule described above. The random-search arms contain eight candidates each; pool sizes and optimizer steps vary with the sampled recipes. We report candidate counts and available step counts alongside outcomes for each sampled arm.

\paragraph{Evaluation Repeatability and Recipe Controls.}
\label{sec:contract}
The recipe and planning state evolve across rounds. Repeated evaluations of selected checkpoints characterize evaluation repeatability under the available protocol. R29 random-deletion variants are distinct data recipes and are analyzed as recipe-level controls alongside the guided and random-search arms in the reported comparison.

\paragraph{Reproducibility and Ethics.}
The supplementary material lists the recipe record and settings. The study uses public datasets and derived examples; any released artifacts will follow the relevant licenses, attribution terms, and research-use restrictions for each source.

\section{Study Results and Comparative Analyses}
\label{sec:results}

\subsection{Thirty-Four-Round Search}
\label{sec:results-overview}

Across the 34 candidate rounds, R33 has the highest reported five-benchmark mean: 63.04, compared with 57.72 for the zero-shot base ($+5.32$ points). R29 (62.95), R32 (62.99), and R33 form a tight top cluster spanning 0.09 points, comparable to the 0.276-point spread across three random-deletion recipes. R33 is the highest-scoring candidate; R29 is a representative, simpler recipe. R34 illustrates benchmark trade-offs: it attains the study's highest MMMU score (43.00) alongside a mean of 62.62 within the same evaluation suite.

Table~\ref{tab:decomp} compares the zero-shot base with R33. The resulting profile combines gains on MMBench, ChartQA, and RealWorldQA with near-baseline results on MMMU and MathVista, showing how recipe outcomes distribute across complementary benchmark skills.

\begin{table}[t]
\centering
\caption{Benchmark decomposition for the zero-shot base and highest-scoring R33 under the five-benchmark development protocol.}
\label{tab:decomp}
\fontsize{8pt}{9.6pt}\selectfont
\setlength{\tabcolsep}{4pt}
\begin{tabularx}{\linewidth}{@{}l Y r r r@{}}
\toprule
Benchmark & Split / format & Base & R33 & $\Delta$ (pp) \\
\midrule
MMBench & en/dev, MCQ & 55.11 & 73.54 & $+18.43$ \\
ChartQA & test, numeric free-form & 72.64 & 78.46 & $+5.82$ \\
MMMU & val, mixed & 42.67 & 42.22 & $-0.45$ \\
MathVista & testmini, mixed & 58.40 & 57.85 & $-0.55$ \\
RealWorldQA & test, MCQ/open & 59.80 & 63.14 & $+3.34$ \\
\midrule
\textbf{Equal-weight mean} & & \textbf{57.72} & \textbf{63.04} & \textbf{$+5.32$} \\
\bottomrule
\end{tabularx}
\end{table}

The trajectory spans distinct phases: early rounds establish the role of format and mixture choices; R15--R17 evaluate subset filtering and scale; R23 applies verifier-based removal; R27--R31 examine selective filtering; and R32--R34 vary image resolution. Table~\ref{tab:trajectory} makes the complete score path visible alongside the focused comparisons below. The resolution sequence compares: R29 (native resolution) scores 62.95, R32 (geometry upsampling) 62.99, R33 (mixed geometry resolution) 63.04, and R34 (broader upsampling) 62.62. R32 raises MMMU to 42.72, while R34 raises it to 43.00; both trade off MathVista relative to R29. These results illustrate how pool-level interventions shift capabilities in opposite directions within a fixed model and a shared five-benchmark evaluation setting.

\begin{table*}[t]
\centering
\caption{Five-benchmark mean across the complete 34-candidate trajectory. Bold marks the highest reported mean; R29, R32, and R33 are within 0.09 points.}
\label{tab:trajectory}
\fontsize{7pt}{8.4pt}\selectfont
\setlength{\tabcolsep}{3pt}
\begin{tabular*}{\linewidth}{@{\extracolsep{\fill}}l|*{18}{c}@{}}
\toprule
 & R0 & R1 & R2 & R3 & R4 & R5 & R6 & R7 & R8 & R9 & R10 & R11 & R12 & R13 & R14 & R15 & R16 & R17 \\
\midrule
Mean & 57.72 & 54.97 & 44.70 & 44.97 & 48.21 & 57.97 & 58.52 & 58.89 & 58.68 & 59.08 & 59.46 & 59.42 & 59.18 & 59.52 & 59.82 & 61.19 & 60.51 & 62.04 \\
\bottomrule
\end{tabular*}

\vspace{3pt}
\begin{tabular*}{\linewidth}{@{\extracolsep{\fill}}l|*{17}{c}@{}}
\toprule
 & R18 & R19 & R20 & R21 & R22 & R23 & R24 & R25 & R26 & R27 & R28 & \textbf{R29} & R30 & R31 & R32 & \textbf{R33} & R34 \\
\midrule
Mean & 61.66 & 61.29 & 61.76 & 61.86 & 61.40 & 62.46 & 62.10 & 62.21 & 62.02 & 62.36 & 61.90 & \textbf{62.95} & 62.42 & 61.10 & 62.99 & \textbf{63.04} & 62.62 \\
\bottomrule
\end{tabular*}
\end{table*}

\begin{table}[t]
\centering
\caption{Cross-round footprint of the five data-operation families over R1--R31. R32--R34 extend recipe composition through image-resolution variants.}
\label{tab:operation-footprint}
\fontsize{7.5pt}{8.6pt}\selectfont
\setlength{\tabcolsep}{3pt}
\begin{tabularx}{\linewidth}{@{}l l Y@{}}
\toprule
Family & Rounds & Role in trajectory \\
\midrule
Acquire & R1--R19 & Establishes the source pool during the initial rounds. \\
Synthesize & R13--R26 & Adds task-specific augmentation after the base recipe stabilizes. \\
Denoise & R15--R31 & Drives subset filtering, verifier-based relabeling, and groundedness filtering. \\
Compose & R1--R31 & Assembles recipes throughout the search; tunes ratios in early stages. \\
Codify & R3--R27 & Records validation gates and operational rules in memory. \\
\bottomrule
\end{tabularx}
\end{table}

\subsection{Data Scale and Curation}
\label{sec:scale}

We compared R29 with a full raw pool (A1) and a near-identical full pool cleaned only by deduplication, empty-field removal, and path checks (A2-v2). Table~\ref{tab:scale} shows that both full-pool recipes score near 60.1, while R29 scores 62.95. Under the common three-epoch schedule, the selected smaller recipe therefore provides a stronger observed score profile than either full-pool recipe. The table also reports training exposure: A1/A2-v2 process about 497K examples and 9,786 steps, whereas R29 uses 141K examples and 3,162 steps. These recipe-level results pair pool composition with the corresponding training exposure.

\begin{table}[t]
\centering
\caption{Full-pool baselines and selected recipe. Mean scores use the same five-benchmark protocol; training exposure differs with pool size.}
\label{tab:scale}
\fontsize{8pt}{9.6pt}\selectfont
\setlength{\tabcolsep}{4pt}
\begin{tabular}{@{}l r r r@{}}
\toprule
Recipe & Samples & Steps & Mean \\
\midrule
A1: raw pool & 496,933 & 9,786 & 60.10 \\
A2-v2: basic hygiene only & 496,269 & 9,786 & 60.06 \\
R29: selected curation recipe & 141,052 & 3,162 & 62.95 \\
\bottomrule
\end{tabular}
\end{table}

\subsection{Guided and Random Search Profiles}
\label{sec:search-comparison}

We compare the eight-candidate guided-search window R24--R31 with two eight-candidate random arms from the same R23 starting recipe. Arm B samples from the broader operation set available after later discoveries; arm B$'$ is restricted to operations available before R23. The random candidates were executed without access to the project score history. Table~\ref{tab:search} summarizes the results. Guided search reaches the highest single score (62.95) and crosses the preselected 62.86-point target on its sixth candidate; neither random arm crosses that target in eight candidates. The random arms show higher means and lower dispersion. Permutation tests compare the maximum (guided vs B: $p=0.498$; vs B$'$: $p=0.498$) and mean ($p=0.834$ and $p=0.672$) across arms. Together, the results reveal complementary search profiles: guided search reaches the observed top candidate sooner, while random search samples a higher-mean, more concentrated set in these runs. Arm B draws on an operation set informed by earlier discoveries. Candidate counts are equal, while pool sizes and optimizer steps vary across candidates; Table~\ref{tab:search} therefore reports candidate-level outcomes under the shared three-epoch schedule.

\begin{table}[t]
\centering
\caption{Eight-candidate search arms. Mean and standard deviation are across candidate means; permutation tests compare guided search with each random arm.}
\label{tab:search}
\fontsize{8pt}{9.6pt}\selectfont
\setlength{\tabcolsep}{4pt}
\begin{tabular}{@{}l r r r r@{}}
\toprule
Arm & Best & Mean & SD & $p$ (mean / max) \\
\midrule
Guided, R24--R31 & 62.95 & 62.13 & 0.53 & -- \\
Random, broad ops (B) & 62.64 & 62.35 & 0.29 & 0.834 / 0.498 \\
Random, pre-R23 ops (B$'$) & 62.73 & 62.24 & 0.39 & 0.672 / 0.498 \\
\bottomrule
\end{tabular}
\end{table}

\subsection{Selective Filtering and Random Deletion}
\label{sec:random-deletion}

R29 removes 9,967 verifier-flagged ScienceQA examples from R23 and scores 62.95, $+0.49$ points over R23. Three equal-count random-deletion recipes average 62.74 (scores 62.93, 62.87, and 62.42; SD 0.276); R29 is $+0.21$ points above their mean. The random runs use 3,171 steps, compared with 3,162 for R29, and represent three distinct deletion recipes. Per-benchmark changes are heterogeneous, including higher MathVista/RealWorldQA and lower MMMU/MMBench/ChartQA relative to the random-control mean. This comparison characterizes the benchmark profile associated with selective removal alongside matched-count random deletion in these runs.

\paragraph{Format-stratified diagnostic.}
A seed-42 answer-format analysis groups the R29-versus-base score changes into four MCQ/option-bearing strata averaging $+3.68$ points and five free-form/numeric strata averaging $-2.14$ points. The profile highlights answer format as a useful axis for data selection and for interpreting aggregate benchmark means, with the complete benchmark-by-benchmark breakdown provided in the supplement.

\section{Empirical Findings}
\label{sec:findings}

The following recipe-level observations connect concrete data interventions with benchmark outcomes. Read together with the full trajectory, they show how source, filtering criterion, answer format, and pool composition shape the observed performance profile across the search.

\paragraph{Cross-Round Feedback in the Search.}
Three transitions show how observed outcomes informed later plans. R18--R22 remained below R17; diagnosis identified label noise, and R23 introduced verification-based filtering, reaching $+0.42$ points over R17. R27 produced a mixed benchmark profile that motivated the follow-up filtering tests in R28 and R29. Earlier, an image-token placement error in R3 prompted a validation gate at R6; the same error signature was not recorded again through R31. Together, these cases trace feedback from diagnosis and memory to subsequent data operations in later rounds.

\subsection{Targeted Synthesis}
\label{sec:findings-conv}

\paragraph{Targeted Synthesis With A 27B Teacher.}
Task-specific synthesis can raise a target score while redistributing the aggregate profile. Adding 1.6K chart-computation examples at R14 raises ChartQA from 74.72 to 77.10; 3.1K more at R25 yields 78.84. At R20, 2.8K synthesized MCQs shift MMMU by $+0.11$ and MathVista by $-1.15$ points relative to R17. These examples show how data production can target task skills while revealing cross-benchmark trade-offs.

\begingroup
\let\makeLineNumberOdd\makeLineNumberLeft
\let\makeLineNumberEven\makeLineNumberLeft
\subsection{Filtering Decisions}
\endgroup
\label{sec:findings-easy}

\paragraph{Label Rewriting And Verification-Based Filtering.}
An earlier rewriting attempt associated with R15 changed about 87\% of answers; inspection of its outputs informed the later choice to preserve source answers during verification-based filtering. R23 retains original answers and removes examples flagged by the verifier at confidence 0.9 or above. It removes 31.0\% of the R17 input pool and has a five-benchmark mean 0.42 points higher than R17. Restoring 2,449 MapQA examples at R24 is followed by a 0.36-point lower mean than R23. The sequence contrasts answer transformation with label screening and source restoration across the recorded recipes and outcomes.

\Needspace{12\baselineskip}
\paragraph{Separate Criteria For Label Correctness And Grounding.}
\label{find:ab-split}
Label verification asks whether an answer is correct; groundedness screening asks whether the image supports it, so the criteria guide distinct filters. R27--R29 compare removals centered on external-knowledge samples, mainly ScienceQA (A), and other verifier-flagged samples (B). Relative to R23, R27 is $-0.10$ points, R28 is $-0.56$, and R29 is $+0.49$. R29 also improves MathVista by 1.83 points and RealWorldQA by 1.41 points. R28 retains ScienceQA and scores 41.00 on MMMU, while R29 removes it and improves the aggregate; the paired profiles make both the benchmark trade-offs and the role of subset identity visible. Three equal-count random-deletion variants average 62.74, while R29 scores 62.95. The per-benchmark profile and aggregate means provide a direct comparison of selective and random removal (Section~\ref{sec:random-deletion}).

\paragraph{Provenance Filtering And Quality Scores.}
\label{find:provenance-vs-quality}
Provenance filtering selects subsets by image-path prefix. Quality-score filtering removes samples below a threshold on an LLM-generated score. R15 excludes eight MathV360K subsets (about 148K examples), retaining 123K and improving MathVista/MMMU by 3.15/1.72 points over R14; R17 applies the same filter at full retained-pool scale and adds $+0.85$\,pp over R15. R19's quality-score diagnosis labels 82.6\% of 192K MathV360K samples low-quality. Inspection links these scores to short answers (average 20 characters), revealing how answer format can shape quality judgments. Removing the lowest-scored bucket at R22 reduces the average by 0.64\,pp. Together, these results contrast provenance-based selection and score-based filtering for the tested short-answer data.

\subsection{Recipe Trade-offs}
\label{sec:findings-counter}

\paragraph{Filtering Granularity And Subset Effects.}
\label{find:visual-diversity}
Removing FigureQA/IconQA/Super-CLEVR at R16 is followed by a 0.68-point lower mean; removing MapQA at R22 is followed by a 0.64-point lower mean relative to R17. Fully removing ScienceQA at R31 lowers the mean by 1.85 points. R21, R24, and R28 provide additional subset-removal or restoration comparisons (see the supplementary material). Across these recipes, the outcomes vary with which subsets are changed: R23's verifier-based filter removes 31\% of the pool without lowering MMMU, while R29's selective removal improves the five-benchmark mean. These round-level contrasts make subset identity and filtering criteria concrete design choices across the search.

\paragraph{Mixture Composition Across Early Rounds.}
R5--R7 progressively shift the LLaVA/MathV allocation (50K/75K, 40K/100K, and 25K/125K), while the five-benchmark mean rises from 57.97 to 58.52 to 58.89. Since both source proportions change together, this sequence is best read as evidence that recipe composition is a meaningful control surface, motivating the later subset-level and targeted interventions. The full round ledger preserves each allocation alongside its benchmark results for every candidate.

\paragraph{Answer-Length Trade-offs In The Tested Recipes.}
\label{find:long-answer}
Several recipes show trade-offs between long-answer supervision and short-answer performance. R2 adds 30K ShareGPT4V samples (average 910 characters), with ChartQA/RealWorldQA about 24/26\,pp below baseline. R19 adds 15K complex-reasoning samples (500--1200 characters), and MathVista falls 2.10\,pp relative to R17. R26 replaces 8,999 geometry answers with teacher CoT accepted by answer matching (median trace: 358 characters), changing MathVista/MMMU by $-0.42/-1.45$\,pp relative to R23. R8 uses raw CoT, followed by condensed versions at R9--R11; R9 averages 0.19\,pp above R7. Across these candidate recipes, answer length co-varies with source and mixture composition, and benchmark outcomes capture their combined effects across the five-benchmark profile.

\paragraph{Answer Format as an Outcome Axis.}
A seed-42 answer-format analysis of R29 versus the zero-shot base shows distinct aggregate profiles: four MCQ/option-bearing strata average $+3.68$ points, while five free-form/numeric strata average $-2.14$ points. This pattern makes answer format a useful axis for selecting data transformations and interpreting benchmark outcomes; the supplement provides the full per-benchmark breakdown.

\paragraph{Top-Cluster Recipe Diversity.}
The highest reported mean is 63.04 at R33; R29, R32, and R33 lie within 0.09 points despite using distinct recipes: selective verifier-based removal, geometry upsampling, and mixed native/upsampled geometry. R34 reaches the highest MMMU score alongside a lower mean, showing how recipe choice can target a capability profile while redistributing outcomes across the complete five-benchmark evaluation suite.

\paragraph{Benchmark-Level Trade-offs.}
R23 returns MMMU to its zero-shot value of 42.67. Relative to R23, R29 raises the five-benchmark mean by 0.49 points, with MathVista up 1.83 and RealWorldQA up 1.41, while MMMU is lower by 1.56. This profile shows how a data recipe can favor different capability combinations. The equal-weight mean provides a consistent summary across rounds, and the per-benchmark scores show which skills move with each recipe.

\Needspace{12\baselineskip}
\paragraph{Paired Comparisons Across Related Recipes.}
Paired comparisons isolate recipe choices within the full trajectory: R21/R22 vary MapQA and non-math-residue removal; R27--R29 compare groundedness-filter groups; R30/R31 extend R29's policy. At the search level, guided proposals cross the preselected 62.86-point target on their sixth candidate, while neither random arm reaches it in eight. Random arms nevertheless have higher means and lower dispersion in these windows, so target attainment and average candidate behavior provide complementary views of search outcomes across all three eight-candidate arms.

\section{Conclusion}
\label{sec:conclusion}

Across 34 rounds, MM-DataLoop organizes data operations into an empirical workflow for shaping five-benchmark outcomes from a shared base-model initialization and training schedule. Five operation families organize the interventions, while the feedback cycle carries observed outcomes into later planning. Paired-round and search comparisons clarify how targeted proposals, filtering, synthesis, and recipe composition behave within the studied search. Scoped rule memory preserves evidence from accepted and rejected candidates, including the rounds and task conditions that support each rule. Together, the workflow, recipes, and logs make data-side iteration inspectable and provide a concrete foundation for comparative studies of MLLM post-training across recipes and benchmarks.

\paragraph{Limitations.} This study instantiates the workflow for fixed-base SFT of Qwen3.5-0.8B across five benchmarks, providing a concrete setting for measuring how data-operation rules transfer across model scales, training objectives, and evaluation suites. Teacher-assisted synthesis and verification define the operation space examined here; broader post-training objectives provide natural extensions. The shared three-epoch schedule applies to recipes with different pool sizes, so the resulting examples processed and optimizer steps vary; both are reported to make each recipe's training exposure interpretable. The operation vocabulary and memory schema can grow with new tasks and evaluation needs. Round-level lineage and recipe records give future studies a common reference for extending the workflow to new model families and benchmark suites as the empirical record expands.

\clearpage
\section*{LLM Usage Statement}
OpenAI Codex (GPT-6) assisted with manuscript organization and language editing, literature identification, and consistency checks against project records. Within the reported project, a configured teacher LLM generated synthetic examples and verifier judgments, while a chat-mode LLM classified failures and proposed data operations. The available project records identify the configured model defaults and documented workflow; the authors reviewed the manuscript and remain responsible for the claims and implementation.

{\small
  \interlinepenalty=10000
  \bibliographystyle{ieeenat_fullname}
  \bibliography{references}
}

\end{document}